\documentclass[%
 reprint,
 amsmath,amssymb,
 aip,
]{revtex4-2}

\usepackage{graphicx}% Include figure files
\usepackage{dcolumn}% Align table columns on decimal point
\usepackage{bm}% bold math
\begin{document}

%\preprint{APS/123-QED}

\title{Design of the Artificial: lessons from the biological roots of general intelligence}% Force line breaks with \\

\author{Nima Dehghani}
%\altaffiliation[This work started initially at: ]{Department of Physics, MIT, Cambridge, MA}%Lines break automatically or can be forced with \\
 \email{nima.dehghani@mit.edu}
\affiliation{%
 McGovern Institute for Brain Research, MIT, Cambridge, MA
}%
%\affiliation{Center for Brains, Minds, and Machines, McGovern Institute for Brain Research, MIT, Cambridge, MA}
\affiliation{Department of Physics, MIT, Cambridge, MA}
 %\homepage{http://www.scholar.harvard.edu/nima}

\date{\today}% It is always \today, today,

%%%% Abstract text to be placed here %%%%%%%%%%%%
\begin{abstract}
Our fascination with intelligent machines goes back to ancient times with the mythical automaton Talos, Aristotle’s mode of mechanical thought (syllogism) and Heron of Alexandria’s mechanical machines. However, the quest for Artificial General Intelligence (AGI) has been troubled with repeated failures. Recently, there has been a shift towards bio-inspired software and hardware, but their singular design focus makes them inefficient in achieving AGI. Which set of requirements have to be met in the design of AGI? What are the limits in the design of the artificial? A careful examination of computation in biological systems suggests that evolutionary tinkering of contextual processing of information enabled by a hierarchical architecture is key to building AGI.
\end{abstract}
%%%%%%%%%%%%%%%%%%%%%%%%%%%

%%%%%%%%%%%%%%% End of first page %%%%%%%%%%%%%%%%%%%%%

\maketitle

% \n{testing Nima's edit} \\
% \m{testing X's edit} \\
% \st{testing deletions} \\ 

\section*{Overview}
The works of late Renaissance and early 17th century visionaries, such as Thomas Hobbes’ mechanical combinatorial theory of cognition in Leviathan, Blaise Pascal’s mechanical calculator and Gottfried Leibniz’s alphabet of human thought, inspired the dream of building intelligent machines. Subsequent centuries saw more serious efforts to develop theories of intelligence and construct intelligent machines. Charles Babbage’s unfinished programmable mechanical calculating machines in late 19th century paved the way for the development of formal logic, Turing’s view of computation and intelligence. The field of Artificial Intelligence emerged in the 20th century, with its ups and downs, and its remarkable achievements. IBM’s deep blue victory over Gary Kasparov and Google DeepMind AlphaGo’s victory over Lee Sedol in the game of GO were among the milestones, though they used strategies very different from human intelligence. Additionally, recent advances in experimental neuroscience and computing power has led to much excitation about the further extension in applications of neuro-inspired computer vision \cite{Cox2014}. However, our machines and algorithms still lack Artificial General Intelligence (AGI). The most advanced AI algorithms are far from the dramatic portrayal of AGI, Hal9000, in 2001 Space Odyssey.

The quest for AGI has been elusive so far. Why did symbolic computation, which seemed promising in the beginning \cite{Newell1961}, fail to achieve AGI? Was it because of the lack of a general database of commonsense knowledge? \cite{McCarthy1987}. Neural networks, which faced early criticism, have proven to be useful in many tasks after the increase of computing power. The magic seems to have been the computing power and the size of the network. But still, these networks are far from AGI. If intelligence is not just about building a synthetic brain \cite{Ball2016}, what is the nature of intelligence? We propose a new perspective that defines intelligence as a mechanism that is closely related to the core essence of biology, i.e. a specially designed hierarchical architecture that enables contextual information processing.

\section*{Promises and limitations so far}
The development of intelligent machines has been accompanied by a noticeable shift towards bio-inspired designs in recent years. In designing the \href{https://youtu.be/dEbLeuUIaHI}{RoboBee} \cite{Ma2013} or the \href{https://youtu.be/-D_XrRo0h20}{RoboRay} \cite{Park2016}
, which mimic the flight of insect or the combination of biological and electronic elements, the key has been to understand the physical limits of the materials used in making flexible lightweight systems. The effectiveness of these systems mostly depends on the communication capacity among the electronics (or bio-electric interface) and processing power of their elements. In more direct attempts to mimic intelligence, the design of novel neuromorphic computing architectures has mainly aimed at achieving brain-like energy efficiency in spiking neural networks embedded designs \cite{Merolla2014,Furber2016}. On the algorithmic front, the successful deep learning \cite{LeCun2015} and recurrent network \cite{Hopfield1982}loose mimicry of brain networks are mainly focused on intrinsic organization of the biological network.

The premise is that such mimicry implies intelligence even though many other elements in physical imitation of biological systems are absent. What is lacking from the Von-Neumann architecture and the bio-inspired non-Von-Neumann approach is an understanding of the computational limits that result from the organization of biological templates as adaptive complex systems embedded in their environment. This oversight is not very surprising since the designers of (both software and hardware) machines focus on the functional aspects of a living system in their attempt to replicate its performance. In the behavioral and functional dichotomy of intelligence, as proposed by \cite{Rosenblueth1943}, the functional approach is concerned with the intrinsic organization while the relation between the object and the outside world is relatively secondary to the structure and internal properties. The emphasis on functional aspects and reliance on high speed components has high energetic costs. Most efficient neural networks, especially the more bio-inspired ones like Deep Learning, need many computational cycles to reach a desired outcome. This also adds to the energetic cost of computation and part of the new efforts in the bio-inspired frontier (neuromorphic architecture) are aimed at resolving this issue. 

The idea of controlling behavior through negative feedback as the mechanism behind teleological behavior \cite{Rosenblueth1943} led to the notion that artificial intelligence and building a synthetic brain are not the same. Feedback as a control mechanism for achieving adaptation in a changing environment unraveled the mysterious aspect of teleology \cite{Simon1962,Simon1969} . Interestingly, early attempts at designing intelligent machines based on teleological behavior showed some promising results \cite{Walter1950}. Walter Grey’s tortoises, \textit{Machina speculatrix} and \textit{Machina docilis} exhibited complex sets of behaviors (such as phototaxis, search for energy source and conditioned reflex behavior) and could navigate around the room autonomously. The teleological approach, though properly recognizing the limits of functional approach, eventually made cybernetics stagnant at a qualitative level \cite{Ball2016}. The reason that neither functional nor behavioral teleology have succeeded in advancing artificial intelligence beyond task-specific achievements is likely rooted in the hierarchical organization of the nervous systems and their modification and optimization through interactions with the environment. It is possible that what distinguishes biology from other systems is the contextual (functional) algorithmic features in biological computation. The context-dependency of biological systems can be defined as a top-down causal information exchange that is unique to their multi-scale organization \cite{Walker2012}. While the thermodynamics of information tells us that information is physical \cite{Landauer1991} and within the bounds of Landauer limit \cite{Landauer1961,Berut2012} could be converted to energy \cite{Parrondo2015,Toyabe2010}, the information at the macroscopic scale can affect microscopic scale in biological systems \cite{Walker2012}.

It has been proposed that the emergence of aggregates over microscopic elements may potentially indicate macro-scale causally superseding micro-scale \cite{Hoel2013}. If macroscopic dynamics can change causality direction then there is the possibility of a strong causal emergence at multiple scales of an information processing system such as the brain. Top-down causality provides a possible platform for transferring the computational complexity onto cells \cite{Pezzulo2016}. One can conclude that through ``information to energy conversion'' and ``top-down supervenience’’ goal-directed macroscopic contextual information processing can reconfigure microscopic dynamics. Of course, in such an information-theoretic landscape, a successful model should tell us how to convert macroscopic goals into computational rules embedded at the scale of networks (neural assemblies) and individual neurons (with potential extension to intracellular computation). The evolved structure that provides the additional top-down causal path (and thus a bidirectional causality), renders information not as mere metaphor of the degree of randomness (as framed by Shannon \cite{Shannon1948}) but gives it a semantic flavor \cite{Smith2000}. This notion of information processing in biological systems is not just a symbolic analogy of modern-day technology (such as Descartes’ hydraulic pump analogy, Freud’s steam engine analogy and the popular ``brain is a computer’’ analogy), but rather it truly reflects the physical nature of information \cite{Smith2000}.%,Seoane2018}.

\section*{Contextual computation}
If the nature of intelligence is rooted in a contextual, non-local information control and feedback, then it follows that there is no fixed solution and no correspondence between the causality of information and microscopic dynamics of the nervous system. It is important to note that this claim does not negate the role of the physical substrate (molecules, neurons and networks) of information flow, but rather depicts the non-optimality and non-uniqueness of the multiscale contextual causality of information processing. In such a setting, the need for heuristics in exposure to new environments/problems becomes evident. ``Trial and error’’ as the basic heuristic strategy derived from experience with similar problems \cite{Pearl1984} and a method of problem solving and error elimination under various forms of selective pressure \cite{Radnitzky1989} stands as the likely candidate. Unsurprisingly, it has been suggested that chess masters do not rely on flash of insight or superior memory or faster processing, but rather implement a heuristic selective and non-exhaustive trial and error in the tree of possible moves \cite{Simon1962a}. Since the non-optimality and non-uniqueness of trial-error seem to be the essence of AGI as well as byproducts of contextual information processing, one may ask where and how they could be better than the alternatives? 

Ashby provides an interesting example of the advantage of a trial and error approach \cite{Ashby1960}. In his example of finding a particular combination of 1000 on/off switches, a simultaneous (parallel) individual testing of all switches (average of 1 second) outperforms serial and all-or-none test of switches (respectively, average of $5x10^2$ and $10^{301}$ seconds). As the number of the possible states of the problem increases, the time needed for any intelligent design to find the optimal solution increases non-linearly. In contrast, trial and error which relies on little knowledge and is problem-specific seems to be the optimal path to reach a solution even though that solution may not be optimal. This contextual, non-local feedback causal aspect of intelligence is very similar to the DNA serving as the backbone but not blueprint of the cellular computation. In the case of DNA, it is the interaction of the genes in response to the cell’s environment that leads to the expression or suppression of other genes \cite{Ball2016,Walker2012}. It is the interactions of the genes and not their products, i.e. enzymes, that carry the information that realize one possibility among many possibilities based on the genetic composition \cite{Smith2000}. The reason for such isomorphism between genetics, immune system and intelligence is the nature of information itself. The allure of the biology is not in its search for the optimal solution but rather is rooted in its fundamental roots in trial and error as a tinkering mechanism in response to the environment \cite{Jacob1977}. As a result, simply by adapting the evolutionary principles, robots can show forms of intelligent behavior \cite{Pfeifer2005} that are systematically different from the mainstream ones. Although for a system to be fully capable of AGI, one has to go above and beyond evolutionary robotics and implement the principles mentioned here.

Brains are (physical) computational systems with a dynamical broken symmetry of information that finds relatively stable states \cite{Hopfield1994}. But how does a system which exhibits context-dependent behavior remain stable? The answer lies in Ashby’s law of requisite variety, which states that ``If a system is to be stable, the number of states of its control mechanism must be greater than or equal to the number of states in the system being controlled’’ \cite{Ashby1958}. As a result, systems with higher internal variety are more capable of responding to predictable scenarios and are better prepared to face unpredictable situations. To achieve a higher number of behaviors, the system needs to increase its intrinsic variety. To do so, the system must rely on a higher number of computing components and a higher number of connections between these computing elements. However, there is a cost of increased connectedness among the components. Up to some level of increased connectedness, the dynamics of the system is stable. However, at some critical threshold, it suddenly becomes unstable \cite{Gardner1970}. If the big system is organized in a modular fashion, it is more likely that individual modules/blocks maintain their stability in the face of perturbations \cite{May1972}, though there is a limit to the number of random interactions and they should be fixed in time \cite{Cohen1985}. This desired stable collective information processing can be achieved through competitive dynamics \cite{Daniels2016} where modular composition provides the possibility of asynchronous or synchronous recruitment of sub-assemblies. It follows that as the system’s variety increases, so does the need for the organization of information and the capacity to reshape the structure as information increases. Evolution has found a solution for this and that solution resides in multiscale organization of biological systems.

Simple life forms are capable of basic computation equipping them with simple behaviors such as chemotaxis and movement. In more complex life forms, especially multicellular animals, behaviors are controlled by the nervous system. Primitive nervous systems (such as jellyfish) are in the form of a diffuse nerve net. As the organism’s behaviors get more complicated, the nervous system becomes organized in ways to adapt to the greater repertoire of the needed behaviors. Comparative structures of the nervous system in C. elegans, fruit fly, mouse, macaque monkey and humans show increasing levels of internal variety, organized in a hierarchical fashion. Ashby’s law of requisite variety requires a higher internal variety to match the more complicated sets of behaviors. Yet, the needed bigger nervous system is not just composed of a very large neuronal net with shallow depth. Instead, the more advanced nervous system is structured as a hierarchical and compositional system. These advanced systems are still composed of microscopic elements involved in the similar types of basic intracellular computation (as in single cell organisms) as well as computations that are built upon a network of cells organized as tissue. As we discussed above, this form of architecture helps to provide dynamic stability of the system while enabling a much larger repertoire of computations.

A natural consequence of a hierarchic structure is that increasing network depth may provide an efficient way for achieving abstraction of information through renormalization \cite{Lin2017}. This aspect may represent why and how deep learning has been more successful than its predecessors in certain domains. Note that the more complex the nervous systems is, the more hierarchical organization is present. Such non-decomposable hierarchy is a hallmark of complex systems \cite{Simon1962,Simon1969}. It is essential to recognize that this hierarchical architecture of complexity provides the opportunity for repair, modification and improvement of different parts of the system without the need to stop the operation of the system. Simon’s parable of the two watchmakers Tempus and Hora \cite{Simon1962, Simon1969}, and their distinctive systems of shallow and serial vs modular and hierarchical approach to watch assembly, reflects how the compositionality of hierarchical complex system can give them a significant edge. Similarly, not only does the hierarchical nature of advanced nervous systems provide a better chance for the repair, modification and improvement of different parts, but also these different subassemblies can be recruited to parallel computational tasks and thus the space of compositional computations greatly increase. Note that evolutionary tinkering also benefits from such hierarchical modularity. These modules can be reused in the re-design of the species’ next generation or can be borrowed and implemented in the construct of other species \cite{Jacob1977}. Whether we study the nature of computation in the nervous system or we wish to design robust AGI, these principles will be crucial.

Note that this notion of intelligence directly opposes all attempts trying to formalize intelligence as an inductive process. Bayesian inference advocating for strong generalization from few examples relies on inference from inference over hierarchical generative models \cite{Tenenbaum2011}. Because of the reasons mentioned above, bayesian formalism is simply unable to achieve the desired (non-optimal) generality of intelligence. Note that the trial and error is probably the most fundamental element of all knowledge gathering systems. In fact from an epistemological point of view, even the core of scientific discovery is not based on induction \cite{Popper1959} \footnote{see Popper's discussions on the logic of scientific discovery}. In contrast to bayesian approach, the complex intelligent behavior has roots in non-symbolic information transfer and arises as a result of interaction of the system with the environment. 

\section*{The road to AGI}
In the design of AGI, the emphasis should be on behavior-oriented Artificial intelligence emerging adaptive intelligence in a constantly changing environment \cite{Steels1993}. Only in such case the relation between scales and the bidirectional causality in biological computation finds a proper meaning. It follows that the context-dependency entails a specific intrinsic multiscale organization of information and provides the ability to interact with the changes in the environment. Thus the limit of an intelligent system is shaped by a core element of complex systems, i.e. multiscale organization, the requisite variety and trial and error. Surprisingly, in formalizing a theory of intelligent systems, these were not considered together before. In an influential essay, \href{https://www.theatlantic.com/magazine/archive/1945/07/as-we-may-think/303881/}{``As We May Think’'}, Vannevar Bush envisioned a future where computers support humans in many different activities \cite{Bush45}. If we truly intend to reach such stage, bio-inspired design will be our solution but only if we take into account the contextual aspects of cognition and intelligence. Evolution has been very successful in creating AGI through trial and error. That blueprint is in front of our nose and ours to mimic.

%\section*{Acknowledgment}
%Insert the Acknowledgment text here.

%\clearpage
% Bibliography
\section*{References}
\bibliographystyle{plain}
\bibliography{biblio}

\end{document}